\documentclass[10pt,twocolumn,letterpaper]{article}

\usepackage{iccv}
\usepackage{times}
\usepackage{epsfig}
\usepackage{graphicx}
\usepackage{amsmath}
\usepackage{amssymb}
\usepackage{subcaption}

\usepackage[breaklinks=true,bookmarks=false]{hyperref}

\iccvfinalcopy % *** Uncomment this line for the final submission

\def\iccvPaperID{****} % *** Enter the ICCV Paper ID here

\ificcvfinal\fi

\begin{document}
	
	\title{The Instantaneous Accuracy: a Novel Metric for the Problem of Online Human Behaviour Recognition in Untrimmed Videos}

	\author{Marcos Baptista R\'ios \\
			University of Alcal\'a \\
			{\tt\small marcos.baptista@uah.es} \and
			Roberto J. L\'opez-Sastre \\
			University of Alcal\'a \\
			{\tt\small robertoj.lopez@uah.es} \and
			Fabian Caba Heilbron \\
			Adobe Research \\
			{\tt\small caba@adobe.com} \and
			Jan van Gemert \\
			Delft University of Technology \\
			{\tt\small j.c.vangemert@tudelft.nl} \and
			F. Javier Acevedo-Rodr\'iguez \\
			University of Alcal\'a \\
			{\tt\small javier.acevedo@uah.es} \and
			Saturnino Maldonado-Basc\'on \\
			University of Alcal\'a \\
			{\tt\small saturnino.maldonado@uah.es}}

	\maketitle
	\ificcvfinal\thispagestyle{empty}\fi

	%%%%%%%%% ABSTRACT
	\begin{abstract}
		The problem of Online Human Behaviour Recognition in untrimmed videos, aka Online Action Detection (OAD), needs to be revisited. Unlike traditional offline action detection approaches, where the evaluation metrics are clear and well established, in the OAD setting we find few works and no consensus on the evaluation protocols to be used. In this paper we introduce a novel online metric, the Instantaneous Accuracy ($IA$), that exhibits an \emph{online} nature, solving most of the limitations of the previous (offline) metrics. We conduct a thorough experimental evaluation on TVSeries dataset, comparing the performance of various baseline methods to the state of the art. Our results confirm the problems of previous evaluation protocols, and suggest that an IA-based protocol is more adequate to the online scenario for human behaviour understanding. Code of the metric available \href{https://github.com/gramuah/ia}{here} (https://github.com/gramuah/ia).
	\end{abstract}

	%%%%%%%%% BODY TEXT
	\section{Introduction}
	\label{sec:introduction}
	We focus on recognizing human behaviours in untrimmed videos \emph{as soon as} they happen, which was coined as Online Action Detection (OAD) by De Geest \etal \cite{DeGeest2016eccv}. 
	
	The problem of action detection has been widely studied, but \emph{mainly from an offline perspective}, \eg \cite{Shou2016cvpr, Yeung2016cvpr, Shou2017cvpr, Zhao2017iccv, Dai2017iccv, Xu2017iccv, Gao2017bmvc_a, Buch2017cvpr, Buch2017bmvc, Chao2018cvpr}, where it is assumed that all the video is available to make predictions. Few works address the \emph{online} setting, \eg \cite{DeGeest2016eccv, Gao2017bmvc_b, DeGeest2018wacv}. Think of a robotic platform that must interact with humans in a realistic scenario, recognizing their behaviours. \emph{All} previous offline methods make this application impossible as they will detect the action situations way later they have occurred. In contrast, OAD approaches must give detections over video streams, hence working with partial observations. However, among the online approaches there is an important weakness in a fundamental aspect: the evaluation metric. We have noticed that there is no consensus on the evaluation protocols, \ie in each dataset a different metric is proposed for the same problem. Moreover, used metrics cannot be said to be of an \emph{online nature}. In other words, the proposed metrics for recent \emph{online} action detection models, such as the mean Average Precision (mAP) or the Calibrated Average Precision (cAP) \cite{DeGeest2016eccv}, do not provide information about the instantaneous performance of the solutions over time: they need to be computed entirely offline, accessing the whole set of action annotations for a given test video.
	
	\begin{figure}[t]
		\begin{center}
			\includegraphics[width=\columnwidth]{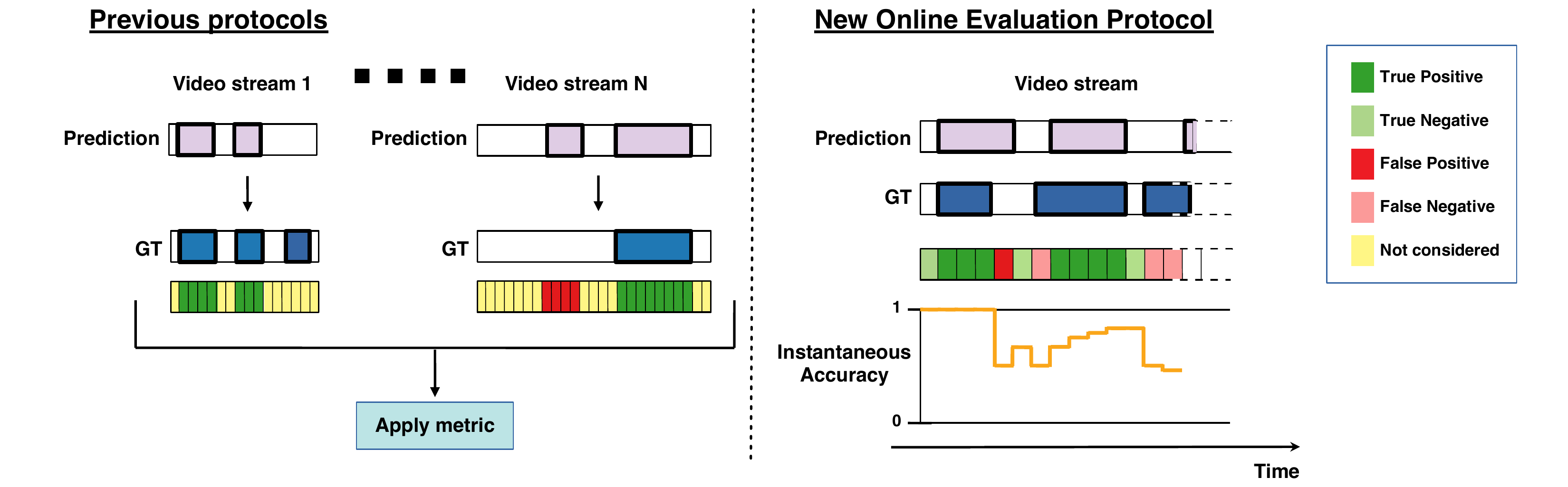}
		\end{center}
		\caption{Previous evaluation protocols for OAD were based on: 1) running the online methods through all videos; 2) applying the offline metric on the obtained results. With our new Instantaneous Accuracy metric (IA), approaches are evaluated online, considering the background, and regardless of the length of the video.}  
		\label{fig:graphical_abstract}
	\end{figure}

	We introduce here an evaluation protocol with a novel \emph{online} metric: the Instantaneous Accuracy ($IA$) (see Figure \ref{fig:graphical_abstract}). This metric has been designed not only to overcome the described limitations, but to allow fair comparisons between OAD methods. A thorough experimental evaluation is conducted on the challenging TVSeries dataset \cite{DeGeest2016eccv}, offering a comparison between baselines and state-of-the-art approaches. Results show that an $IA$-based evaluation protocol is more adequate for the OAD problem, because it is able to give a detailed evolution of the performance of OAD models when the video stream grows over time.
	
	%%%%%%%%% INSTANTANEOUS ACCURACY
	\section{The Instantaneous Accuracy}
	\label{sec:ia}
	We argue that an evaluation protocol for OAD must evaluate method performances as the video grows over time with an online video-level metric.
	
	\noindent\textbf{Previous metrics.} All previous evaluation protocols use class-level metrics which have to be applied offline, \ie at the end of the test time. These protocols are mainly based on the per-frame mean average precision (mAP) or its calibrated version (cAP) \cite{DeGeest2016eccv}.
	
	\noindent\textbf{Instantaneous Accuracy metric.} Considering a set of $ \mathcal{N} $ test videos, for each video, an OAD method generates a set of action detections defined by their initial and ending times. IA metric takes as input these detections to build a dense temporal prediction of action (including background) for every time slot $ \Delta t $ in the test video. Note $ \Delta  t$ is the unique parameter of our IA metric and it measures how often the metric is computed. For a particular instant of time $ t' $, the $ \mathrm{IA}(t') $ is computed as the time slot-level accuracy for the action classification task as follows:
	\begin{equation}
	\mathrm{IA}(t') = \frac{\sum_{j=0:\Delta t:t'}\vec{tp}(j) + \sum_{j=0:\Delta t:t'}\vec{tn}(j)}{K'} \, ,
	\label{eq:ia}
	\end{equation}
	where $ \vec{tp} $ and $ \vec{tn} $ are two vectors encoding the true positives (actions) and true negatives (background), respectively, and $ K' $ represents the total population considered until time $ t' $, which is dynamically obtained as follows: $ K' = \left\lfloor \left( \frac{t'}{\Delta t} \right) \right\rfloor\, . $
	
	As working with untrimmed videos, where much more background than action frames appear, we propose a weighted version of the IA. Technically, we simply scale in Eq. \ref{eq:ia} the \emph{true} factors by the dynamic ratio between background and action slots until time $t'$ in the ground truth.
	
	To summarise method's performance across a dataset for research purposes, we propose to use the mean average Instantaneous Accuracy ($\mathrm{maIA}$) for every video:
	\begin{equation}
	\mathrm{maIA} = \frac{1}{N} \sum_{i=1:N} \left( \frac{\Delta t}{T_i} \sum_{j=0:\Delta t:T_i} \mathrm{IA}(j) \right) \, . 
	\end{equation}
	
	%%%%%%% EXPERIMENTS
	\section{Experiments and conclusions}
	\label{sec:experiments}
	
	We use the challenging TVSeries dataset \cite{DeGeest2016eccv}, following the setup in \cite{Gao2017bmvc_b} to analyse all the metrics considered in our study: mAP, cAP, and the novel online IA.
	
	As baselines, we propose the following: All background (\textbf{All-BG}), which simply simulates a model that never generates an action class; Perfect Model (\textbf{PM}), that always assigns correct labels to action and background frames and helps to reveal the limitations of the previous evaluation protocols, showing their metrics cannot saturate to the maximum which they have been designed for; and finally, we propose a \textbf{3D-CNN} model, which consists in a C3D network \cite{DuTran2015iccv} to recognise all actions and the background category.
	
	\begin{table}[t!]
		\centering
		\caption{Analysis of the metrics on TVSeries.}
		\resizebox{7cm}{!}{
			\begin{tabular}{c|c|ccc}
				& CNN \cite{DeGeest2016eccv} & All-BG  & 3D-CNN & PM  \\ \hline
				mAP (\%) & 1.9 & 0 & 1.6 & 30.9 \\ 
				cAP (\%) & 60.8 & 0 & 10.8 & 96.9 \\ \hline
				maIA (\%) & 3.51 & 78.3 & 71.9 & 100 \\
				weighted maIA (\%) & 12.46 & 22.9 & 28.9 & 100 \\ \hline
			\end{tabular}
		}
		\label{tab:tvseries}
	\end{table}
	
	Table \ref{tab:tvseries} shows the results for these baselines and state-of-the-art model in \cite{DeGeest2016eccv}. First, one observes that offline protocols do not succeed in giving a 100\%, even for a perfect method. This is due to their incorrect way of managing the background category. The calibrated AP seems to alleviate this problem, but it still does not achieve a 100\%. Second, results from All-BG baseline reveal the relevance of having a weighted metric, especially for an imbalanced problem. And third, 3D-CNN achieves competitive performance when compared to the state of the art, supporting its choice as a strong baseline for the OAD problem. It is only for the cAP where its performance decreases compared with CNN~\cite{DeGeest2016eccv}, but the reason is that CNN does not cast predictions for background category (while 3D-CNN does), and the cAP has been designed to minimize the importance of such errors.
	
	\begin{figure}[h!]
		\begin{subfigure}[b]{\columnwidth}
			\centering
			\includegraphics[width=.8\columnwidth]{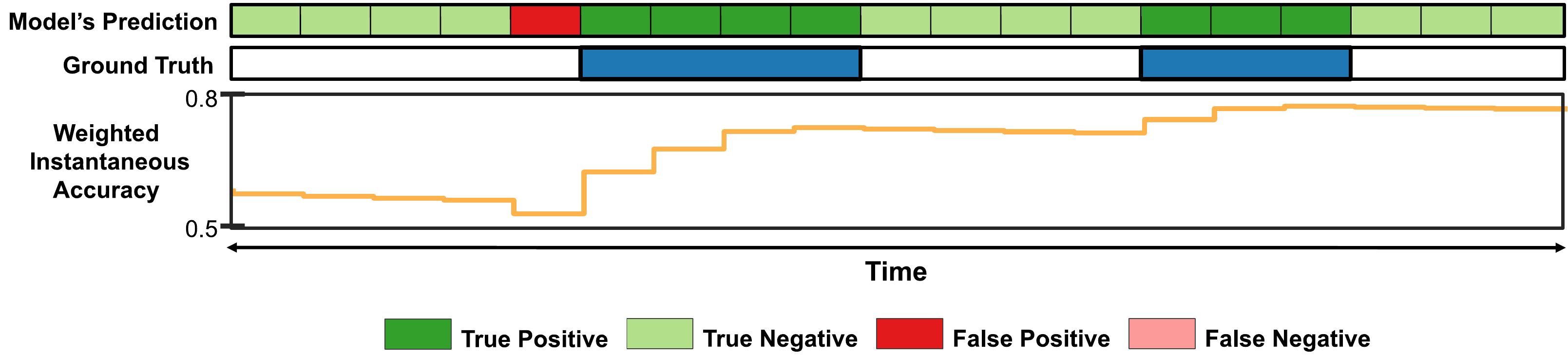}
			\caption{}
			\label{subfig:qualitative-results}
		\end{subfigure}
		\begin{subfigure}[b]{\columnwidth}
			\centering
			\includegraphics[width=.8\columnwidth]{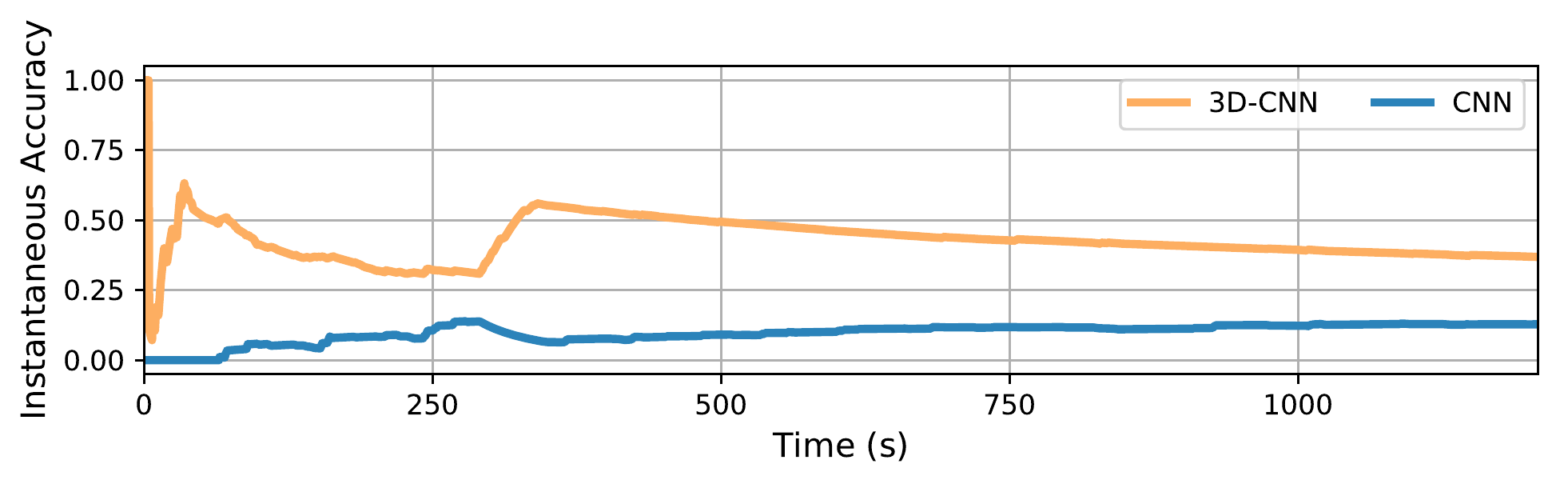}
			\caption{}
			\label{subfig:ia-cnn-c3d}
		\end{subfigure}
		\caption{(a) Evolution of the weighted IA for a video for the 3D-CNN baseline. (b) Weighted IA comparison between 3D-CNN and CNN \cite{DeGeest2016eccv}.}
	\end{figure}
	
	Figure \ref{subfig:qualitative-results} shows the evolution of the weighted $IA$ for a particular video. We use here for visualization 0.5 seconds for $\Delta t$. One can observe how the weighting mechanism works, dynamically adapting the IA to the observed proportion of the video. Figure \ref{subfig:ia-cnn-c3d} also shows a comparison between CNN \cite{DeGeest2016eccv} and our 3D-CNN.
	
	As a conclusion, online human behaviour recognition in untrimmed videos is a challenging task with few contributions. We found limitations in the metrics used so far, so we have introduced a new online metric that complies with the online nature of the problem: the Instantaneous Accuracy ($IA$). Experimental results have proved both the limitations of previous used metrics and the robustness of IA.
	
{\small
\bibliographystyle{ieee_fullname}
\bibliography{egbib}

\begin{thebibliography}{10}\itemsep=-1pt

\bibitem{Buch2017cvpr}
S. {Buch}, V. {Escorcia}, C. {Shen}, B. {Ghanem}, and J.~C. {Niebles}.
\newblock Sst: Single-stream temporal action proposals.
\newblock In {\em IEEE Conference on Computer Vision and Pattern Recognition
  (CVPR)}, pages 6373--6382, July 2017.

\bibitem{Chao2018cvpr}
Y. {Chao}, S. {Vijayanarasimhan}, B. {Seybold}, D.~A. {Ross}, J. {Deng}, and R.
  {Sukthankar}.
\newblock Rethinking the faster r-cnn architecture for temporal action
  localization.
\newblock In {\em IEEE Conference on Computer Vision and Pattern Recognition
  (CVPR)}, pages 1130--1139, June 2018.

\bibitem{Dai2017iccv}
X. {Dai}, B. {Singh}, G. {Zhang}, L.~S. {Davis}, and Y.~Q. {Chen}.
\newblock Temporal context network for activity localization in videos.
\newblock In {\em IEEE International Conference on Computer Vision (ICCV)},
  pages 5727--5736, October 2017.

\bibitem{DeGeest2016eccv}
Roeland De~Geest, Efstratios Gavves, Amir Ghodrati, Zhenyang Li, Cees Snoek,
  and Tinne Tuytelaars.
\newblock Online action detection.
\newblock In {\em European Conference on Computer Vision (ECCV)}, pages
  269--284, October 2016.

\bibitem{DeGeest2018wacv}
R. {De Geest} and T. {Tuytelaars}.
\newblock Modeling temporal structure with lstm for online action detection.
\newblock In {\em IEEE Winter Conference on Applications of Computer Vision
  (WACV)}, pages 1549--1557, March 2018.

\bibitem{Gao2017bmvc_a}
Jiyang Gao, Zhenheng Yang, and Ram Nevatia.
\newblock Cascaded boundary regression for temporal action detection.
\newblock In {\em British Machine Vision Conference (BMVC)}, pages 521--5211,
  September 2017.

\bibitem{Gao2017bmvc_b}
Jiyang Gao, Zhenheng Yang, and Ram Nevatia.
\newblock {RED:} reinforced encoder-decoder networks for action anticipation.
\newblock In {\em British Machine Vision Conference (BMVC)}, pages 921--9211,
  September 2017.

\bibitem{Shou2017cvpr}
Z. {Shou}, J. {Chan}, A. {Zareian}, K. {Miyazawa}, and S. {Chang}.
\newblock Cdc: Convolutional-de-convolutional networks for precise temporal
  action localization in untrimmed videos.
\newblock In {\em IEEE Conference on Computer Vision and Pattern Recognition
  (CVPR)}, pages 1417--1426, July 2017.

\bibitem{Shou2016cvpr}
Z. {Shou}, D. {Wang}, and S. {Chang}.
\newblock Temporal action localization in untrimmed videos via multi-stage
  cnns.
\newblock In {\em IEEE Conference on Computer Vision and Pattern Recognition
  (CVPR)}, pages 1049--1058, June 2016.

\bibitem{Buch2017bmvc}
Bernard~Ghanem Shyamal~Buch, Victor~Escorcia and Juan~Carlos Niebles.
\newblock End-to-end, single-stream temporal action detection in untrimmed
  videos.
\newblock In {\em British Machine Vision Conference ({BMVC})}, pages 931--9312,
  September 2017.

\bibitem{DuTran2015iccv}
D. {Tran}, L. {Bourdev}, R. {Fergus}, L. {Torresani}, and M. {Paluri}.
\newblock Learning spatiotemporal features with 3d convolutional networks.
\newblock In {\em IEEE International Conference on Computer Vision (ICCV)},
  pages 4489--4497, Dec 2015.

\bibitem{Xu2017iccv}
H. {Xu}, A. {Das}, and K. {Saenko}.
\newblock R-c3d: Region convolutional 3d network for temporal activity
  detection.
\newblock In {\em IEEE International Conference on Computer Vision (ICCV)},
  pages 5794--5803, October 2017.

\bibitem{Yeung2016cvpr}
S. {Yeung}, O. {Russakovsky}, G. {Mori}, and L. {Fei-Fei}.
\newblock End-to-end learning of action detection from frame glimpses in
  videos.
\newblock In {\em IEEE Conference on Computer Vision and Pattern Recognition
  (CVPR)}, pages 2678--2687, June 2016.

\bibitem{Zhao2017iccv}
Y. {Zhao}, Y. {Xiong}, L. {Wang}, Z. {Wu}, X. {Tang}, and D. {Lin}.
\newblock Temporal action detection with structured segment networks.
\newblock In {\em IEEE International Conference on Computer Vision (ICCV)},
  pages 2933--2942, October 2017.

\end{thebibliography}
}

\end{document}